\documentclass[twoside,11pt]{article}

\usepackage{blindtext}

\usepackage{jmlr2e}

\usepackage{lastpage}
\jmlrheading{24}{2023}{1-\pageref{LastPage}}{1/23; Revised 9/23}{12/23}{23-0064}{Zongyi Li,  Daniel Zhengyu Huang,  Burigede Liu,  Anima Anandkumar}
\ShortHeadings{Geometric Fourier Neural Operator}{Li, Huang, Liu, Anandkumar}

\firstpageno{1}

\usepackage{natbib}

\usepackage[utf8]{inputenc} 
\usepackage[T1]{fontenc}    
\usepackage{hyperref}       
\usepackage{url}            
\usepackage{booktabs}       
\usepackage{amsfonts}       
\usepackage{nicefrac}       
\usepackage{microtype}      
\usepackage[dvipsnames]{xcolor}         

\usepackage{graphicx}
\usepackage{enumitem}
\usepackage{amsmath}
\usepackage{mathtools}
\usepackage{stfloats}
\usepackage{bm}
\usepackage{cleveref}
\newcommand{\kamyar}[1]{}
\newcommand{\zongyi}[1]{}
\newcommand{\aacomment}[1]{}

\newcommand{\N}{\mathbb{N}}

\newcommand{\R}{\mathbb{R}}

\newcommand{\E}{\mathbb{E}}

\newcommand{\cK}{\mathcal{K}}
\newcommand{\cP}{\mathcal{P}}
\newcommand{\cQ}{\mathcal{Q}}

\newcommand{\F}{\mathcal{F}}
\newcommand{\G}{\mathcal{G}}
\newcommand{\Gtrue}{\mathcal{G}^{\dagger}}

\newcommand{\T}{\mathcal{T}}

\newcommand{\I}{\mathbb{I}}

\usepackage{enumitem} 
\begin{document}

\title{Fourier Neural Operator with Learned Deformations\\ 
for PDEs on General Geometries}

%

\author{\name Zongyi Li
\email zongyili@caltech.edu \\
\addr Computing and Mathematical Science\\
California Institute of Technology\\
Pasadena, CA 91125, USA
\AND 
\name Daniel Zhengyu Huang 
\email huangdz@bicmr.pku.edu.cn \\
\addr Beijing International Center for Mathematical Research\\
Peking University\\
Beijing, 100871, China
\AND 
\name Burigede Liu 
\email bl377@cam.ac.uk \\
\addr Department of Engineering \\
University of Cambridge\\
Cambridge, CB2 1PZ, UK
\AND 
\name Anima Anandkumar 
\email anima@caltech.edu \\
\addr Computing and Mathematical Science\\
California Institute of Technology\\
Pasadena, CA 91125, USA
}


\editor{Lorenzo Rosasc}

\maketitle

\begin{abstract}
Deep learning surrogate models have shown promise in solving partial differential equations (PDEs). Among them, the Fourier neural operator (FNO) achieves good accuracy, and is significantly faster compared to numerical solvers,  on a variety of   PDEs, such as fluid flows. However, the FNO uses the Fast Fourier transform  (FFT), which is limited to rectangular domains with uniform grids. In this work, we propose a new framework, viz., Geo-FNO, to solve PDEs on arbitrary geometries. Geo-FNO learns to deform the input (physical) domain, which may be irregular, into a latent space with a uniform grid. The FNO model with the FFT is applied in the latent space. The resulting Geo-FNO model has both the computation efficiency of FFT and the flexibility of handling arbitrary geometries. Our Geo-FNO is also flexible in terms of its input formats, viz.,  point clouds, meshes, and design parameters are all valid inputs. We consider a variety of PDEs such as the Elasticity, Plasticity, Euler's, and Navier-Stokes equations, and both forward modeling and inverse design problems. Comprehensive cost-accuracy experiments show that Geo-FNO is $10^5$ times faster than the standard numerical solvers and twice more accurate compared to direct interpolation on existing ML-based PDE solvers such as the standard FNO.

\end{abstract}



\section{Introduction}
Data-driven engineering design has the potential to accelerate the design process by orders of magnitude compared to conventional methods. It can enable extensive exploration of the design space and yield new designs with far greater efficiency. This is because the conventional design process requires repeated evaluations of partial differential equations~(PDEs) for optimization, which can be time-consuming. Examples include computational fluid dynamics (CFD) based aerodynamic design and topology optimization for additive manufacturing. 
Deep learning approaches have shown promise in speeding up PDE evaluations and automatically computing derivatives, hence accelerating the overall design cycle.
However, most deep learning approaches currently focus on predicting a few important quantities in the design process, e.g. lift and drag in aerodynamics, as opposed to completely emulating the entire simulation (i.e., pressure or Mach number fields in aerodynamics). This limits the design space to be similar to the training data, and may not generalize to new geometries and scenarios. In contrast, we develop an efficient deep learning-based approach for design optimization, which emulates the entire PDE simulation on general geometries leading to a versatile tool for exploring the design space.

\paragraph{Deformable meshes.}
Solutions of PDEs frequently have high variations across the problem domain, and hence some regions of the domain often require a finer mesh compared to other regions for obtaining accurate solutions. For example, in airfoil flows, the region near the airfoil requires a much higher resolution for accurately modeling the aerodynamics, as shown in Figure \ref{fig:fluid}.
To address such variations, deformable mesh methods such as adaptive finite element methods (FEM) have been developed. Two adaptive mesh methods are commonly used: the adaptive mesh refinement method that adds new nodes (and higher-order polynomials), and the adaptive moving mesh method that relocates the existing nodes~\cite{babuvska1990p,huang2010adaptive}. 
While mesh refinement is popular and easy to use with traditional numerical solvers, it changes the mesh topology and requires special dynamic data structures, which reduce speed and make it hard for parallel processing. 
On the other hand, the adaptive moving mesh methods retain the topology of the mesh, making it possible to integrate with spectral methods. 
Spectral methods solve the equation on the spectral space (i.e., Fourier, Chebyshev, and Bessel), which usually have exponential convergence guarantees when applicable~\cite{gottlieb1977numerical}. 
However, these spectral methods are limited to simple computational domains with uniform grids. When the mesh becomes non-uniform, spectral basis functions are no longer orthogonal and the spectral transform is no longer invertible, so the system in the deformed Fourier space is not equivalent to the original one anymore. Thus, traditional numerical methods are slow on complex geometries due to computational constraints. 

\paragraph{Neural operators.}
Deep learning surrogate models have recently yielded promising results in solving PDEs.  One class of such models is data-driven, and they directly approximate the input-output map through supervised learning. Such models have achieved significant speedups compared to traditional solvers in numerous applications~\citep{Zabaras, bhatnagar2019prediction}. Among them, the graph neural networks have not been studied for complex geometries~\citep{allen2022physical, sanchez2020learning}
Recently, a new class of data-driven models known as neural operators aim to directly learn the solution operator of the PDE in a mesh-convergent manner. Unlike standard deep learning models, such as convolutional neural networks from computer vision, neural operators are consistent to discretization and hence, better suited for solving PDEs. They generalize the previous finite-dimensional neural networks to learn operator mapping between infinite-dimensional function spaces. Examples include Fourier neural operator (FNO)~\citep{li2020neural}, graph neural operator \cite{li2020multipole}, 
and DeepONet \citep{lu2019deeponet}.   We consider FNO~\citep{li2020fourier} in this paper due to its superior cost-accuracy trade-off~\citep{de2022cost}. 
FNO implements a series of layers computing global convolution operators with the fast Fourier transform (FFT) followed by mixing weights in the frequency domain and inverse Fourier transform. These global convolutional operators are interspersed with non-linear transformations such as GeLU. 
By composing global convolution operators and non-linear activations, FNO can approximate highly non-linear and non-local solution operators.
FNO and its variants are able to simulate many PDEs such as the Navier-Stokes equation and seismic waves, do high-resolution weather forecasts, and predict CO2 migration with unprecedented cost-accuracy trade-off~\citep{pathak2022fourcastnet, yang2021seismic, wen2022u}.

\paragraph{Limitations of FNO on irregular geometries.}
While FNO is fast and accurate, it has limitations on the input format and the problem domain. Since FNO is implemented with FFT, it can be only applied on rectangular domains with uniform meshes.  When applying it to irregular domain shapes, previous works usually embed the domain into a larger rectangular domain~\citep{lu2022comprehensive}. However, such embeddings are less efficient and wasteful, especially for highly irregular geometries.
Similarly, if the input data is in the form of non-uniform meshes such as triangular meshes,   previous works use basic interpolation methods between the input non-uniform mesh and a uniform mesh on which FFT is computed~\cite{li2020fourier}. This can cause large interpolation errors, especially for non-linear PDEs. 


\begin{figure}
    \centering
    \includegraphics[width=\textwidth]{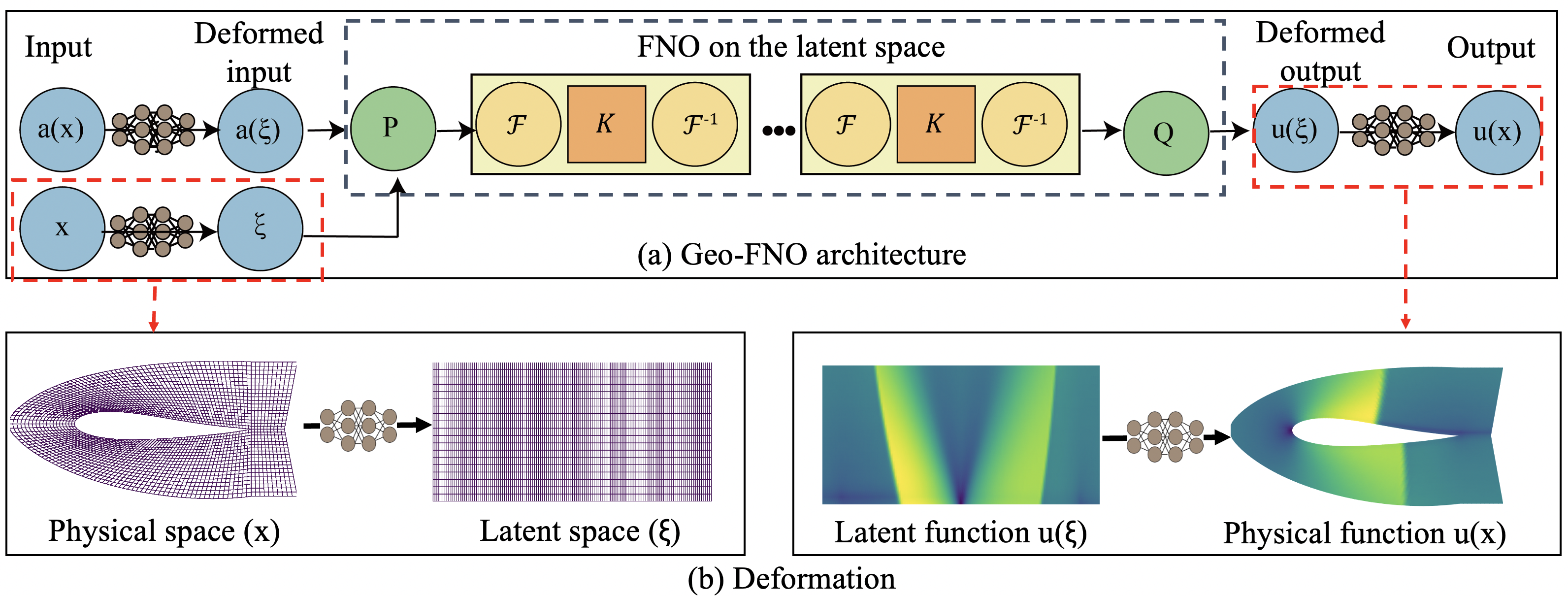}
    \caption{Geometry-aware FNO}
    \label{fig:GFNO}
    {\small  
    \textbf{Top:}  The Geo-FNO model deforms the input functions from irregular physical space to a uniform latent space. After the standard FNO \cite{li2020fourier} is applied, the Geo-FNO will deform the latent solution function back to the physical space. $P$, $Q$ are lifting and projection. $\F$ and $\F^{-1}$ are the Fourier transforms. $K$ is a linear transform  \eqref{eq:G}. 
    \textbf{Bottom:} The deformation, either given or learned,  defines a correspondence between the physical space and computational space. It induces an adaptive mesh and a generalized Fourier basis on the physical space.} 
\end{figure}

\paragraph{Our contributions.}
To address the above challenges, we develop a geometry-aware Fourier neural operator (Geo-FNO) and we summarize our contributions below:
\begin{enumerate}[leftmargin=*]
    \item We propose a geometry-aware discretization-convergent FNO framework (Geo-FNO) that works on arbitrary geometries and a variety of input formats such as point clouds, non-uniform meshes,  and design parameters.
    \item Geo-FNO deforms the irregular input domain into a uniform latent mesh on which the FFT can be applied. Such deformation can be fixed or learned with the FNO architecture in an end-to-end manner. For the latter case, we design a neural network to model the deformation and train it end-to-end with Geo-FNO.
    \item We experiment on different geometries for the Elasticity, Plasticity, Advection, Euler's, and Navier-Stokes equations
    on both forward modeling and inverse design tasks. 
    The cost-accuracy experiments show that Geo-FNO has up to $10^5$  acceleration compared to the numerical solver, as well as half the error rate compared to previous interpolation-based methods as shown in Figure \ref{fig:cost-accuracy}. Further, Geo-FNO maintains discretization-convergence. The model trained on low resolution datasets can be directly evaluated at high resolution cases.
\end{enumerate}

Geo-FNO thus learns a deformable mesh along with the solution operator in an end-to-end manner by applying a series of FFTs and inverse FFTs on a uniform latent mesh, as shown in Figure \ref{fig:GFNO}.   
Thus, Geo-FNO combines both the computational efficiency of the FFT  and the flexibility of learned deformations. It is as fast as the original FNO but can represent the variations in solutions and domain shapes more efficiently and accurately. The learned deformation from the physical irregular domain to a uniform computational domain is inspired by the traditional adaptive moving mesh method \citep{huang2010adaptive}. However, the adaptive moving mesh method has not had much success with traditional numerical solvers due to the loss of orthogonality of the spectral transform. On a deformed mesh, the system in the Fourier space is no longer equivalent to the original system, so it does not make sense to solve the PDE in the deformed Fourier space. 
However, Geo-FNO does not have this limitation. It does not involve solving equations in the Fourier space. Instead, Geo-FNO approximates the solution operator via the computation on Fourier space in a data-driven manner.

In principle, our Geo-FNO framework can be directly extended to general topologies. Given complex input topology we can apply the decomposition techniques such as triangle tessellation to divide the domain into sub-domains such as each sub-domain is regular. Similarly, for other input formats such as the signed distance functions (SDF), we can sample collocation points. Furthermore, it is possible to extend the Geo-FNO framework to the physics-informed neural operator setting which incorporates PDE constraints \citep{li2021physics}. Since the deformation is parameterized by a neural network, we can obtain the exact derivatives and apply the chain rule to compute the residual error.



\section{Problem Settings and Preliminaries}
\paragraph{Problem settings.}
In this work, we consider parametric partial differential equations defined on various domains. Assume the problem domain $D_{a}$ is parameterized by design parameters $a \in \mathcal{A}$, which follows some distribution {$a \sim \mu$}. For simplicity, we assume all the domains are bounded, orientable manifolds embedded in some background Euclidean space $\Omega$ (e.g., $\R^3$). We consider both stationary problems and time-dependent problems of the following forms:
\begin{align}\label{eq:dynamic}
\begin{split}
    \frac{du}{dt} &= \mathcal{R}(u), \qquad  \text{in } D_{a} \times T \\
    u &= u_0, \qquad \qquad \text{in } D_{a} \times \{0\} \\
    u &= u_b, \qquad \quad \:\:\: \text{in } \partial D_{a} \times T 
\end{split}
\end{align}
where \(u_0 \in \mathcal{U}(0)\) is the initial condition;  \(u_b\) is the boundary condition; and \(u(t) \in \mathcal{U}(t)\) for \(t > 0\) is the target solution function. \(\mathcal{R}\) is a possibly non-linear differential operator. We assume that \(u\) exists and is bounded for all time and for every $u_0 \in \mathcal{U}(t)$ at every $t \in T$. This formulation gives rise to the solution operator \(\Gtrue : (a, u_0, u_b) \mapsto u\). Prototypical examples include fluid mechanics problems such as the Burgers' equation and the Navier-Stokes equation and solid mechanics problems such as elasticity and plasticity defined on various domain shapes such as airfoils and pipes. In this paper, we assume the initial condition $u_0$ and the boundary condition $u_b$ are fixed. Specifically, we also consider stationary problem $\mathcal{R}(u) = 0$ where $u_0$ is not necessary. In this case, the solution operator simplifies to \(\Gtrue : a \mapsto u\).

\paragraph{Input formats.}
In practice, the domain shape can be parameterized in various ways. It can be given as meshes, functions, and design parameters.
In the most common cases, the domain space is given as meshes (point clouds) $a = \T = \{x_i\}^{N} \subset \Omega $, such as triangular tessellation meshes used in many traditional numerical solvers. 
Alternatively, the shapes can be described by some functions $a = f: \Omega \to \R$. For example, if the domain is a 2D surface, then it can be parameterized by its boundary function { $f: [0,1]\to \partial D_a$}. Similarly, the domain can be described as signed distance functions (SDF) or occupancy functions. These are common examples used in computer vision and graphics.
At last, we also consider specific design parameters $a \in \R^d$. For example, it can be the width, height, volume, angle, radius, and spline nodes used to model the boundary. The design parameters restrict the domain shape in a relatively smaller subspace of all possible choices. This format is most commonly used in a design problem, which is easy to optimize and manufacture. For the latter two cases, it is possible to sample a mesh $\T$ based on the shape.

\subsection{Learning the solution operator}
Given a PDE as defined in \eqref{eq:dynamic} and the corresponding solution operator \(\Gtrue\), one can use a neural operator ${\G_{\theta}}$ with parameters ${\theta}$ as a surrogate model to approximate \(\Gtrue\). 
Usually we assume a dataset $\{a_j, u_j\}_{j=1}^N$ is available, where $\Gtrue(a_j) = u_j$ and $a_j \sim \mu$ are i.i.d. samples from some distribution \(\mu\) supported on \(\mathcal{A}\). In this case, one can optimize the solution operator by minimizing the { relative} empirical data loss on a given data pair
{
\begin{align}
\label{eq:pinn-data}
\mathcal{L}_{\text{data}}(u, \G_\theta(a)) = \frac{\|u - \G_\theta(a)\|_{\mathcal{L}^2}}{\|u \|_{\mathcal{L}^2}}= \sqrt{ \frac{\int_{D_a} |u(x) - \G_\theta(a)(x) |^2 \mathrm{d}x}{\int_{D_a} u(x)^2 \mathrm{d}x} }
\end{align}
}
The operator data loss is defined as the average error across all possible inputs
{
\begin{align}
\label{eq:op-data}
\begin{split}
\mathcal{J}_{\text{data}}( \G_\theta) = \E_{a \sim \mu}[ \mathcal{L}_{\text{data}}(\G^{\dagger}(a),\G_\theta(a))] \approx \frac{1}{N} \sum_{j=1}^N \sqrt{ \frac{\int_{D_a} |u_j(x) - \G_\theta(a_j)(x) |^2 \mathrm{d}x}{\int_{D_a} u_j(x)^2 \mathrm{d}x} }.
\end{split}
\end{align}
When the prediction $u$ is time-dependent, the $\mathcal{L}^2$ integration should also account for the temporal dimension. 
}


\subsection{Neural operator }

In this work, we will focus on the neural operator model designed for the operator learning problem.
The neural operator, proposed in \citep{li2020neural}, is formulated as a generalization of standard deep neural networks to operator setting. Neural operators compose linear integral operator $\cK$ with pointwise non-linear activation function \(\sigma\) to approximate highly non-linear operators.
\begin{definition}[Neural operator $\G_\theta$] Define the neural operator
\begin{equation}
    \G_{\theta} \coloneqq \cQ \circ(W_{L} + \cK_{L} + b_L) \circ \cdots \circ \sigma(W_1 + \cK_1 + b_1) \circ \cP
\end{equation}
where \(\cP: \R^{d_a} \to \R^{d_{1}}\), \(\cQ: \R^{d_{L}} \to \R^{d_u}\) are the pointwise neural networks that encode the lower dimension function into higher dimensional space and vice versa. The model stack $L$ layers of $\sigma(W_{l} + \cK_{l} + b_{l})$ where \(W_l \in \R^{d_{{l+1}} \times d_{l}}\) are pointwise linear operators (matrices), \(\cK_l: \{D \to \R^{d_{l}}\} \to \{D \to \R^{d_{l+1}}\}\) are integral kernel operators, $b_l: D \to \R^{d_{l+1}} $ are the bias term made of a linear layer, and \(\sigma\) are fixed activation functions. The parameters $\theta$ consists of all the parameters in $\cP, \cQ, W_l, \cK_l, b_l$.
\end{definition}

\begin{definition}[Fourier integral operator $\cK$] Define the Fourier integral operator
\begin{equation}
\label{eq:Fourier}
\bigl(\cK(\phi)v_t\bigr)(x)=   
\F^{-1}\Bigl(R_\phi \cdot (\F v_t) \Bigr)(x) \qquad \forall x \in D 
\end{equation}
where $R_\phi$ is the Fourier transform of a periodic function $\kappa: \bar{D} \to \R^{d_v \times d_v}$ parameterized by \(\phi \in \Theta_\cK\).
\end{definition}
The Fourier transform $\F$ is defined as 
\begin{align}
    (\F v)(k) 
    = \langle v, \psi(k) \rangle_{L(D)}
    = \int_x v(x) \psi(x,k) \mu(x) 
    \approx \sum_{x\in\T} v(x) \psi(x,k)
\end{align}
where $\psi(x,k) = e^{2i \pi \langle x, k \rangle} \in L(D)$ is the Fourier basis and $\T$ is the mesh sampled from the distribution $\mu$.
In \cite{li2020neural}, the domain $D$ is assumed to be a periodic, square torus and the mesh $\T$ is uniform, so the Fourier transform $\F$ can be implemented by the fast Fourier transform algorithm (FFT). In this work, we aim to extend the framework to non-uniform meshes and irregular domains.


\section{Geometry-Aware Fourier Neural Operator }
The idea of the geometry-aware Fourier neural operator is to deform the physical space into a regular computational space so that the fast Fourier transform can be performed on the computational space. 
Formally, we want to find a diffeomorphic deformation $\phi_a$ between the input domain $D_a$ and the unit torus $D^c = [0,1]^d$. 
The computational mesh $D^c$ is shared among all input space $D_a$. It is equipped with a uniform mesh and standard Fourier basis. Once the coordinate map $\phi_a$ is determined, the map induces an adaptive mesh and deformed Fourier basis on the physical space. In the community of numerical PDEs, the diffeomorphism corresponds to the adaptive moving mesh \citep{huang2010adaptive}.

\subsection{Deformation from the physical space to the computational space.}
Let $D_a$ be the physical domain and $D^c$ be the computational domain. Let $x\in D_a$ and $\xi \in D^c$ be the corresponding mesh points. Denote the input meshes as {$\T_a = \{x^{(i)}\} \subset D_a$} and the computational meshes as $\T^c = \{\xi^{(i)}\} \subset D^c$. For simplicity, we assume the output mesh is the same as the input mesh.
The adaptive moving mesh is defined by a coordinate transformation $\phi_a$ that transforms the points from the computational mesh to the physical mesh, as shown in Figure \ref{fig:GFNO} (b)
\begin{equation}
\begin{split}
    \phi_a: D^c &\to D_a\\
     \xi &\mapsto x 
\end{split}
\end{equation}
Ideally, $\phi_a$ is a diffeomorphism, meaning it has an inverse $\phi_a^{-1}$, and both $\phi_a$ and $\phi_a^{-1}$ are smooth. When such a diffeomorphism exists, there is a correspondence between the physical space and the computational space.
Let $\T^c \subset D^c$ be the uniform mesh and $\psi^c \in L(D^c)$ be the standard spectral basis defined on the computational space. The coordinate transformation $\phi_a \subset D_a$ induces an adaptive mesh $\T_a$ and adaptive spectral basis $\psi_a \in L(D_a)$ (pushforward)  
\begin{equation}
\label{eq:deformed-basis}
\begin{split}
    \T_a &\coloneqq \phi_a(\T^c)\\
    \psi_a(x) &\coloneqq \psi^c\circ\phi_a^{-1}(x) 
\end{split}
\end{equation}
It can be interpreted as solving the system $\mathcal{R}(v)=0$ with adaptive mesh $\T_a$ and generalized basis $\psi_a$. Conversely, for any function defined on the physical domain $v \in L(D_a)$, it can be transformed to the computational domain $v^c \in L(D^c)$ (pullback) 
\begin{equation}
\begin{split}
     v^c(\xi) \coloneqq  v(\phi_a(\xi)) 
\end{split}
\end{equation}
Similarly, for any system of equations $\mathcal{R}(v)=0$ such as \eqref{eq:dynamic} defined on the physical domain $D_a$, the transformation $\phi_a$ induced a new deformed system of equations $\mathcal{R}^c(v^c)=0$.
It can be interpreted as solving the deformed system $\mathcal{R}^c(v^c)=0$ with uniform mesh $\T^c$ and standard basis $\psi^c$.

{
\paragraph{Examples: Chebyshev Basis.}
Chebyshev method is a standard spectral method for solving PDE with non-periodic boundary \cite{driscoll2014chebfun}.
It can be induced from the standard cosine basis (in discrete cosine transform) with a cosine grid. The Chebyshev polynomials on domain $[-1, 1]$ has the form of 
\[\phi_{\rm cheb}^k := \cos( k\cdot \cos^{-1}(x)),\quad k= 0, 1, 2 \ldots\]
It's equivalent to apply
 the cosine deformation $\phi_a: [0, \pi] \to [-1, 1]$
 \[\phi_a(x) = \cos(x)\] to the basis of cosine series
\[\phi_{\cos}^k : = \cos(k\cdot x) ,\quad k= 0, 1, 2 \ldots\]
in the same manner as a deformed basis \eqref{eq:deformed-basis} 
\[\phi_{\rm cheb}^k = \phi_{\cos}^k \circ \phi_a^{-1}(x)\]
Intuitively, the cosine deformation places more grid points around the boundary $-1$ and $1$, which addresses the stability issue around the boundary. Notice, in practice the discrete cosine  transform on $[0, \pi]$ can be implemented by the faster Fourier transform using an extended domain $[-\pi, \pi]$ and restricts to real coefficients. We will discuss the use of extended domain in Section \ref{sec:fourier-continuation} Fourier continuation. Therefore, GeoFNO equipped with the cosine deformation can implement the Chebyshev method.  However, it's hard to deform the Fourier basis into spherical basis, since the underlying domain is not homeomorphic. For these non-homeomorphic domain we will use domain decomposition as discussed in Section \ref{sec:domain-decompsition}. 
}

\subsection{Geometric Fourier transform}
Based on the deformation, we can define the spectral transform in the computational space.
Fourier transforms conducted in the computational domain are standard, since we have a uniform structured mesh in the computational domain. In this subsection, We will introduce a geometric spectral transform between the function $v(x)$ in the physical domain $D_a$ and the function $\hat{v}(k)$ in the spectral space of the computation domain $D^c$.

To transform the function $v(x)$ from the physical domain to the spectral space of the computation domain, we define the forward transform:
\begin{align}
    (\F_{a} v)(k) &:= \int_{D^c} v^c(\xi)  e^{- 2i \pi \langle \xi, k\rangle}  d\xi\\
                  &= \int_{D} v(x)  e^{- 2i \pi \langle \phi_{a}^{-1}(x), k\rangle}  \lvert\textrm{det}[ \nabla_x \phi_{a}^{-1}(x)]\rvert dx \\
                  &\approx \frac{1}{|\T_a|}\sum_{x\in \T_a}  m(x) v(x)  e^{- 2i \pi \langle \phi_{a}^{-1}(x), k\rangle} \label{eq:Fa}
\end{align}
where the weight $m(x) = \lvert\textrm{det}[ \nabla_x\phi_{a}^{-1}(x)]\rvert / \rho_a(x)$
and $\rho_a$ is a distribution from which the input mesh $\T_a$ is sampled.
Notice, in many cases, we have an input mesh $\T_a$ so that computational mesh $ \phi_a^{-1}(\T_a) = T^c$ is uniform. In this case, $\lvert\textrm{det}[ \nabla_x\phi_a^{-1}(x)]\rvert = \rho_a(x)$, and $m(x) = 1$ can be omitted. Other choices can be made, for example, we can define
$m(x)$ using heuristics depending on the solution $u(x)$ or residual error. The weight $m(x)$ corresponds to the monitor functions in the literature of adaptive meshes, where the adaptive mesh is finer near $x$,  when $m(x)$ is large.

To transform the spectral function $\hat{v}(k)$ from the spectral space of the computation domain to the physical domain, we define the inverse transform
\begin{align}  \label{eq:iFa}
    (\F^{-1}_{a} \hat{v})(x) = (\F^{-1} \hat{v})(\phi_{a}^{-1}(x)) 
                             = \sum_{k}  \hat{v}(k) e^{2i \pi \langle \phi_{a}^{-1}(x), k \rangle }
\end{align}
It is worth mentioning that $\F_{a} \circ \F^{-1}_{a} = I$, since both are defined on the computational space.
\begin{align}
    (\F_{a} \circ \F^{-1}_{a} \hat{v})(k) &= \int_{D^c} (\F^{-1}_{a} \hat{v})^c (\xi)  e^{- 2i \pi \langle \xi, k\rangle}  d\xi \\
                                          &= \int_{D^c} \sum_{k}  \hat{v}(k) e^{2i \pi \langle \xi, k \rangle } e^{- 2i \pi \langle \xi, k\rangle}  d\xi
                                       \quad   = \hat{v}(k)
\end{align}

Notice both $\F_a$ \eqref{eq:Fa} and $\F^{-1}_a$ \eqref{eq:iFa} only involve the inverse coordinate transform $\phi_{a}^{-1}: D_a \to D^c$. Intuitively, in the forward Fourier transform $\F_a$, we use  $\phi_{a}^{-1}$ to transform the input function to the computational space, while in the inverse Fourier transform $\F^{-1}_a$, we use  $\phi_{a}^{-1}$ to transform the query point to the computational space to evaluate the Fourier basis. It makes the implementation easy in that we only need to define  $\phi_{a}^{-1}$.

\subsection{Architecture of Geo-FNO and the deformation neural network}
We consider two scenarios: (1) the coordinate map is given, and (2) learning a coordinate map. In many cases, the input mesh $\T_a$ is non-uniform but structured. For example, the meshes of airfoils are usually generated in the cylindrical structure. If the input mesh {is structured, meaning it can be indexed as a multi-dimensional array $\T_a[i_1,\ldots,i_d]$ with $0 \leq i_k \leq s_k$ $\forall k$}, then its indexing induces a canonical coordinate map
\begin{equation}
\label{eq:structured}
\phi_a^{-1}: \T_a[i_1,\ldots,i_d] \mapsto (i_1/s_1,\ldots,i_d/s_d)    
\end{equation}
{In this case, $(i_1/s_1,\ldots,i_d/s_d)$ forms a uniform mesh within a unit cube, allowing for the direct application of the FFT. And Geo-FNO reduces to a standard FNO directly applied to the input meshes.}

When we need to learn a coordinate map, we parameterize the coordinate map $\phi_a^{-1}$ as a neural network and learn it along with the solution operator $G_\theta$ in an end-to-end manner with loss \eqref{eq:op-data}. 
{
\begin{equation}
\label{eq:mesh}
\phi_{a}^{-1}: (x_1, x_2, \cdots, x_d, a) \mapsto (\xi_1, \xi_2, \cdots, \xi_d)  
\end{equation}
{We parameterize the deformation as \(\phi_{a}^{-1}(x, a) = x + f(x, a)\), where \(f\) is a standard feed-forward neural network. Specially, $f$ takes input $(x_1,x_2,\cdots, x_d, a)$, where $a$ is the geometry parameters.
}
This formulation initializes $\phi_{a}^{-1}$ around an identity map, which is a good initial choice. 
Following the works of implicit representation \citep{mildenhall2020nerf}, \citep{sitzmann2020implicit}, we use sinusoidal features in the first layers $\sin(B x), B = 2^i$ to improve the expressiveness of the network. 
Concretely, we define $f$ as a three-layer feed-forward neural network with a width of $32$. The input to this network comprises a combination of components, namely $x$ and $a$, alongside a series of trigonometric terms such as $\sin(2^1 x)$, $\cos(2^1 x)$, $\sin(2^2 x)$, $\cos(2^2 x)$, $\cdots, \sin(2^K x)$ and $\cos(2^K x)$, which are applied coordinatewisely. And $a$ can represent the coordinates of the point clouds that encode the design geometry.
Finally, $\phi_a^{-1}$ is used to compute $\F_a$ and $\F_a^{-1}$ (See \eqref{eq:Fa} and \eqref{eq:iFa}).
}

As shown in Figure \ref{fig:GFNO}, Geo-FNO has a similar architecture as the standard FNO, but with a deformation in the input and output
\begin{equation}
\label{eq:G}
    \G_{\theta} \coloneqq \cQ \circ(W_{L} + \cK_{L}(\phi_{a}) + b_L) \circ \cdots \circ \sigma(W_1 + \cK_1(\phi_{a}) + b_1) \circ \cP
\end{equation}
{
Here, we begin by employing $\cP$ to lift (encode) the input to a higher channel dimension. Subsequently, the first Fourier convolution operator $\cK_1$ employs the geometric Fourier transform $ \F_a$ \eqref{eq:Fa}, while the last layer $\cK_L$ employs the inverse geometric Fourier transform $\F^{-1}_a$ \eqref{eq:iFa}. To finalize the process, we utilize $\cQ$ to project (decode) the output back to the desired dimension.
In the case where the input is presented as a structured mesh, we define the coordinates as \eqref{eq:structured}. Under this circumstance, both $\F_a$ and $\F^{-1}_a$ reduce to the standard FFT, leading to Geo-FNO being reduced to the standard FNO.
}

\subsection{Fourier continuation}
\label{sec:fourier-continuation}
For some PDE problems, the input domain has an irregular topology that is not homeomorphic to a disk or a torus. Consequentially, there does not exist a diffeomorphism between $D_a$ and $D^c$. In this case, we will first embed the input domain into a larger regular domain
\[ D_a \xhookrightarrow{i} \bar{D}_a \]
so that $\bar{D}_a$ is diffeomorphic to $D^c$. For example, the elasticity problem \cref{ssec:solid-elastic} has a hole in its square domain. We can first embed the domain into a full square by completing the hole. Such embedding corresponds to the Fourier continuation \citep{bruno2007accurate} techniques used in conventional spectral solvers. Usually it requires to extend the function $u \in \mathcal{U}(D_a)$  to $\bar{u} \in \mathcal{U}(\bar{D}_a)$ by fitting polynomials. However, in the data-driven setting, a such extension can be done implicitly during the training. We only need to compute the loss on the original space $D_a$ and discard the extra output from the underlying space $\bar{D}_a$. 
{This continuation technique is universal. According to the Whitney embedding theorem, any m-dimensional manifold can be smoothly embedded into a Euclidean space $\R^{2m}$.}

\subsection{Domain decomposition}
\label{sec:domain-decompsition}
{For complex topologies, another technique is to decompose the domain into multiple sub-domains such that each of the domains is equipped with a standard topology. We mainly consider 2-dimensional manifolds in our experiments, although the technique is applicable for any dimension. Given any 2-manifold $\mathcal{M}$, we apply the decomposition
$$\mathcal{M} = D_1 \# \cdots \# D_n $$
where $\#$ is the connect-sum, and each $D_i$ is homeomorphic to either a sphere $S^2$ or torus $T^2$. We then train $n$ sub-models $\G_1 \cdots \G_n$, each equipped with a deformation map $\phi_{a_1} \cdots \phi_{a_n}$. This domain decomposition is also universal since any orientable compact 2-dimensional manifold is homeomorphic to a sphere or a $n$-genus torus, where $n$-genus torus can be naturally decomposed into $n$ tori. Compared to continuation, the decomposition method does not require raising the dimensionality, but it needs multiple sub-models. A numerical example is included in Section \ref{ssec:fluid-sphere}. 
}




\begin{figure}
    \centering
    \includegraphics[width=14cm]{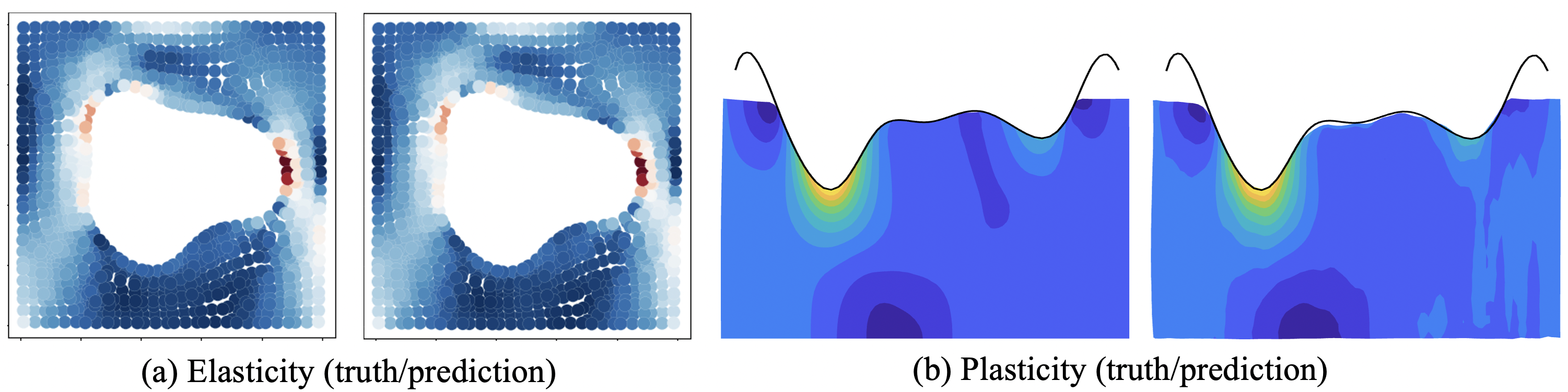}
    \caption{Elasticity (a) and plasticity problems (b) introduced in \cref{ssec:solid-elastic,ssec:solid-plastic}. The comparison is shown between the reference obtained using a traditional solver (left) and the Geo-FNO result (right).}
    \label{fig:solid}
\end{figure}

\begin{figure}
    \centering
    \includegraphics[width=0.6\textwidth]{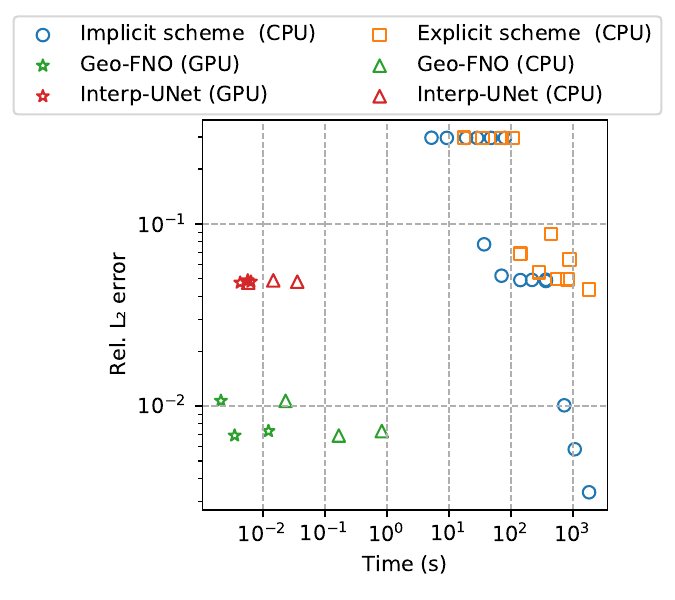}
    \caption{The cost-accuracy trade-off for Geo-FNO/UNet and traditional numerical solver with implicit/explicit temporal integrators on the airfoil problem.}
    \label{fig:cost-accuracy}
\end{figure}

\section{Numerical Examples}
{ In this section, we study Geo-FNO numerically, and compare the Geo-FNO against other machine-learning models on PDEs with various geometries, including hyper-elastic problem in \cref{ssec:solid-elastic}, plastic problem in \cref{ssec:solid-plastic}, advection equation on a sphere problem in \cref{ssec:fluid-sphere}, airfoil problem in \cref{ssec:fluid-airfoil}, and pipe problem in \cref{ssec:fluid-pipe}.
We demonstrate that Geo-FNO can be applied on irregular domains and non-uniform meshes (See \cref{fig:solid,fig:fluid}) and show it is more accurate than combining direct interpolation with existing ML-based PDE solvers such as the standard FNO~\citep{li2020fourier} and UNet\citep{ronneberger2015u}, as well as mesh-free methods such as Graph neural operators (GNO) \citep{li2020neural} and DeepONet \citep{lu2019deeponet} (See \cref{table:elas} and \cref{table:structured}). Meanwhile, Geo-FNO perseveres the speed of standard FNO. It can accelerate the numerical solvers up to $10^5$ times on the airfoil problem (See \cref{ssec:fluid-airfoil}). All experiments are performed on a single Nvidia 3090 GPU. With no special mention, we train all the models with 500 epochs with an initial learning rate of $0.001$ which decays by half every 100 epochs. 
We use the relative $L^2$ error for both the training and testing. 
And the standard Adam optimizer is used.
The code is available at \url{https://github.com/neuraloperator/Geo-FNO}.
}

\subsection{Hyper-elastic problem.} 
\label{ssec:solid-elastic}
The governing equation of a solid body can be written as
\begin{align*}
\rho^s \frac{\partial^2 \bm{u}}{\partial t^2} + \nabla \cdot \bm{\sigma} &= 0
\end{align*}
where $\rho^s$ is the mass density, $\bm{u}$ is the displacement vector, $\bm{\sigma}$ is the stress tensor. Constitutive models, which relate the strain tensor $\bm{\epsilon}$ to the stress tensor, are required to close the system. 
We consider the unit cell problem $\Omega = [0,1]\times[0,1]$ with an arbitrary shape void at the center, which is depicted in Figure \ref{fig:solid} (a). The prior of the void radius is $r = 0.2 + \frac{0.2}{1 + \exp(\tilde{r})}$ with $\tilde{r} \sim \N(0, 4^2(-\nabla + 3^2)^{-1})$, which embeds the constraint $0.2 \leq r \leq 0.4$. 
{The unit cell is clamped on the bottom edge and tension traction $\bm{t} = [0,\,100]$ is applied on the top edge.
The material is the incompressible Rivlin-Saunders material~\cite{pascon2019large} and the constitutive model of the material is given by 
\begin{align*}
\bm{\sigma} &= \frac{\partial w(\bm{\epsilon})}{\partial \bm{\epsilon}}\\
 w(\bm{\epsilon}) &= C_1(I_1 - 3) + C_2(I_2 - 3)
\end{align*}
where $I_1 = \rm{tr}(C)$ and $I_2 = \frac{1}{2}[(\rm{tr}(C)^2 - \rm{tr}(C^2)]$ are scalar invariants of the right Cauchy Green stretch tensor $C = 2\bm{\epsilon}+\I$. And energy density function parameters are $C_1 = 1.863\times10^5$ and $C_1 = 9.79\times10^3$.}
We generate $1000$ training data and $200$ test data with a finite element solver~\cite{huang2020learning} with about $100$ quadratic quadrilateral elements.
It takes about $5$ CPU seconds for each simulation.
The inputs $a$ are given as point clouds with a size of around 1000. The target output is the stress field.

\begin{figure}
    \centering
    \includegraphics[width=14cm]{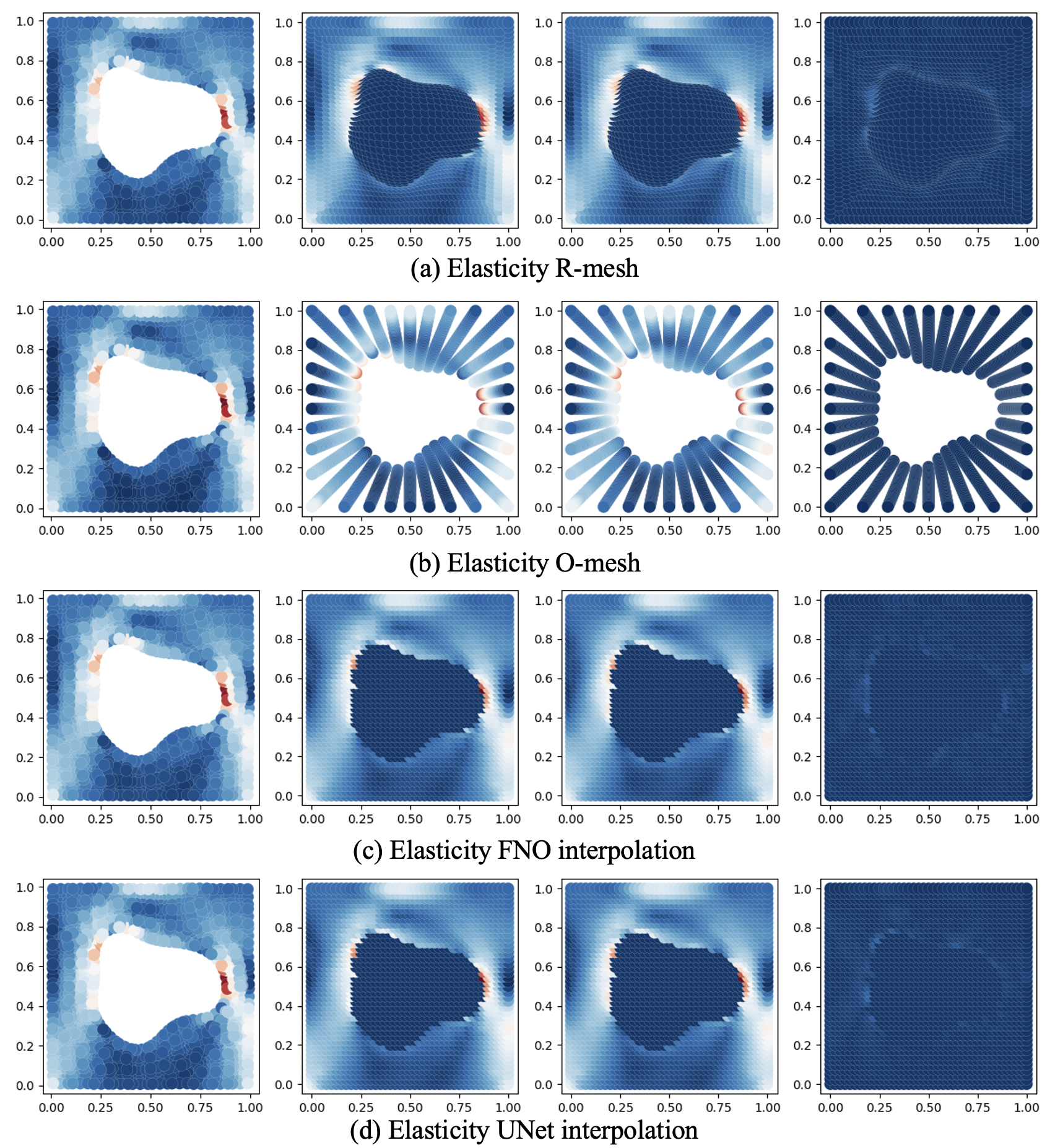}
    \caption{Interpolation into different meshes for the elasticity problem}
    \label{fig:meshes}
    {\small  
    The first column is the original unstructured input; the second column is the interpolated input; the third column is the prediction; the last column is the error. As shown in the figures, interpolation causes an error, which is less accurate than geometry-aware methods.} 
\end{figure}

The elasticity problem has an unstructured input format. We compare the mesh-convergent methods such as Geo-FNO, Graph neural operator (GNO), and Deep operator network (DeepONet), as well as interpolation-based ML methods including uniform meshes (FNO, UNet) and heuristic adaptive meshes (R-mesh, O-mesh). The details of these methods are listed below.
\begin{itemize}
\item GNO: Since the graph structure is flexible on the problem geometry, graph neural networks have been widely used for complex geometries \citep{sanchez2020learning, pfaff2020learning}. In this work, we compare with the graph kernel operator \citep{li2020neural,li2020multipole}, which is a graph neural network-based neural operator model. It implements the linear operator $\mathcal{K}$ by message passing on graphs. We build a full graph with edge connection radius $r=0.2$, width $64$, and kernel with $128$.
\item DeepONet: the deep operator network \citep{lu2019deeponet} is a line of works designed with respect to the linear approximation of operator as shown in \citep{chen1995universal}. It has two neural networks a trunk net and a branch net to represent the basis and coefficients of the operator. Both DeepONets and neural operators aim to parameterize infinitely dimensional solution operators, but neural operators directly output the full field solution functions, while DeepONets output the solution at specific query points. This design difference gives neural operators, especially FNO, an advantage when the data are fully observed while DeepONet has an advantage when the data are partially observed. In this work, we use five layers for both the trunk net and branch net, each with a width of $256$.
\item FNO interpolation: as a baseline, we simply interpolate the input point cloud into a 41-by-41 uniform grid and train a plain FNO model \citep{li2020fourier}. As shown in figure \ref{fig:meshes} (c), the interpolation causes an interpolation error, which loses information on the singularities (the red regions). As a result, the testing error is constantly higher than $5\%$.
\item UNet interpolation: Similarly, we train a UNet model \citep{ronneberger2015u} on the interpolated uniform grid. As FNO-interpolation, the error is constantly higher than $5\%$ as shown in the figure \ref{fig:meshes} (d).
\item Geo-FNO (R mesh): We consider a heuristic adaptive mesh method for each input geometry by deforming a uniform 41 by 41 grid, as shown in the figure
\ref{fig:meshes} (a). The stretching is applied in the radial direction originated at the void center~$(0.5,\,0.5)$ to attain a finer mesh near the interface of the void. Let $r$ denote the radial distance of mesh points, the deformation map is
 \begin{align*}
     r \rightarrow
     \begin{cases}
     r_s  + \alpha(r - r_s) + (1-\alpha)\frac{(r - r_s)^2}{r_e - r_s} & r \geq r_s\\
     r_s  - \alpha(r_s - r) - (1-\alpha)\frac{(r_s - r)^2}{r_s}   & r \leq r_s
     \end{cases}
 \end{align*}
 where $r_s$ and $r_e$ denote the void radius and the unit cell radius in this radial direction, $\alpha=0.2$ denotes the stretching factor, where the ratio of mesh sizes between these near the void interface and these near the unit cell boundary is about $\frac{\alpha}{2-\alpha}$. It is worth mentioning that the deformation map remains void interface, the origin, and the unit cell boundary. Once the adaptive meshes are generated, we interpolate the input data to the adaptive meshes. The surrogate model is learned on the adaptive meshes. We used four Fourier layers with mode $12$ and width $32$.
\item Geo-FNO (O mesh): Similarly, we design another heuristic adaptive mesh method with cylindrical meshing, as shown in the figure \ref{fig:meshes} (b). A body-fitted O-type mesh is generated for each geometry with $64$ points in the azimuth direction and $41$ points in the radial direction.  The points in the azimuth direction are uniformly located between $0$ and $2\pi$, and the points in the radial direction are uniformly located between $r_s$ and $r_e$, where $r_s$ and $r_e$ denote the void radius and the unit cell radius. Again, we interpolate the input data to the adaptive meshes. The surrogate model is learned on the adaptive meshes. We used four Fourier layers with mode $12$ and width $32$.
\end{itemize}

\begin{table*}
\begin{center}
\begin{tabular}{l|crc|cc}
\multicolumn{1}{c}{\bf Model}
&\multicolumn{1}{c}{\bf mesh size}
&\multicolumn{1}{c}{\bf model size}
&\multicolumn{1}{c}{\bf training time}
&\multicolumn{1}{c}{\bf training error }
&\multicolumn{1}{c}{\bf testing error }\\
\hline 
\hline 
Geo-FNO (learned) & 972 & 1546404 & 1s &\textbf{0.0125}& \textbf{0.0229}  \\
GraphNO          & 972 & 571617 & 32s &$0.1271$& $0.1260$\\
DeepONet     & 972 & 1025025 & 45s &$0.0528$& $0.0965$\\
\hline
Geo-FNO (R mesh) & $1681$ & 1188417 & 0.6s &$0.0308$& $0.0536$\\
Geo-FNO (O mesh) & $1353$ & 1188385 & 0.5s &$0.0344$& $0.0363$\\
FNO interpolation & $1681$ & 1188353 & 0.5s &$0.0314$& $0.0508$ \\
UNet interpolation & $1681$ & 7752961 & 0.9s &$0.0089$& $0.0531$ \\ 
\hline
\hline 
\end{tabular}
\end{center}
\caption{Benchmark on the elasticity problem. Inputs are point clouds. { The table presents the mesh size as the number of input mesh points, the model size as the quantity of training parameters, the training time for each epoch, and the relative training and test errors. }} 
\label{table:elas}
\end{table*}

As shown in Table \ref{table:elas}, for the elasticity problem, the Geo-FNO model has a significantly lower error compared to combining direct interpolation with existing ML-based PDE solvers such as the standard FNO\citep{li2020fourier} and UNet\citep{ronneberger2015u}. Both FNO+interpolation and UNet+interpolation methods have a test error larger that $5\%$, which is likely caused by the interpolation error. Geo-FNO also has a lower error compared to mesh-free methods such as Graph neural operators (GNO) \citep{li2020neural} and DeepONet \citep{lu2019deeponet} due to the advantages of the spectral transform,  similar to standard FNO \citep{de2022cost}. Notice that the Geo-FNO with a learned deformation has better accuracy compared to Geo-FNO with {
fixed heuristic deformations (R-mesh) and (O-mesh).}

\begin{figure}[t]
    \centering
    \includegraphics[width=0.8\textwidth]{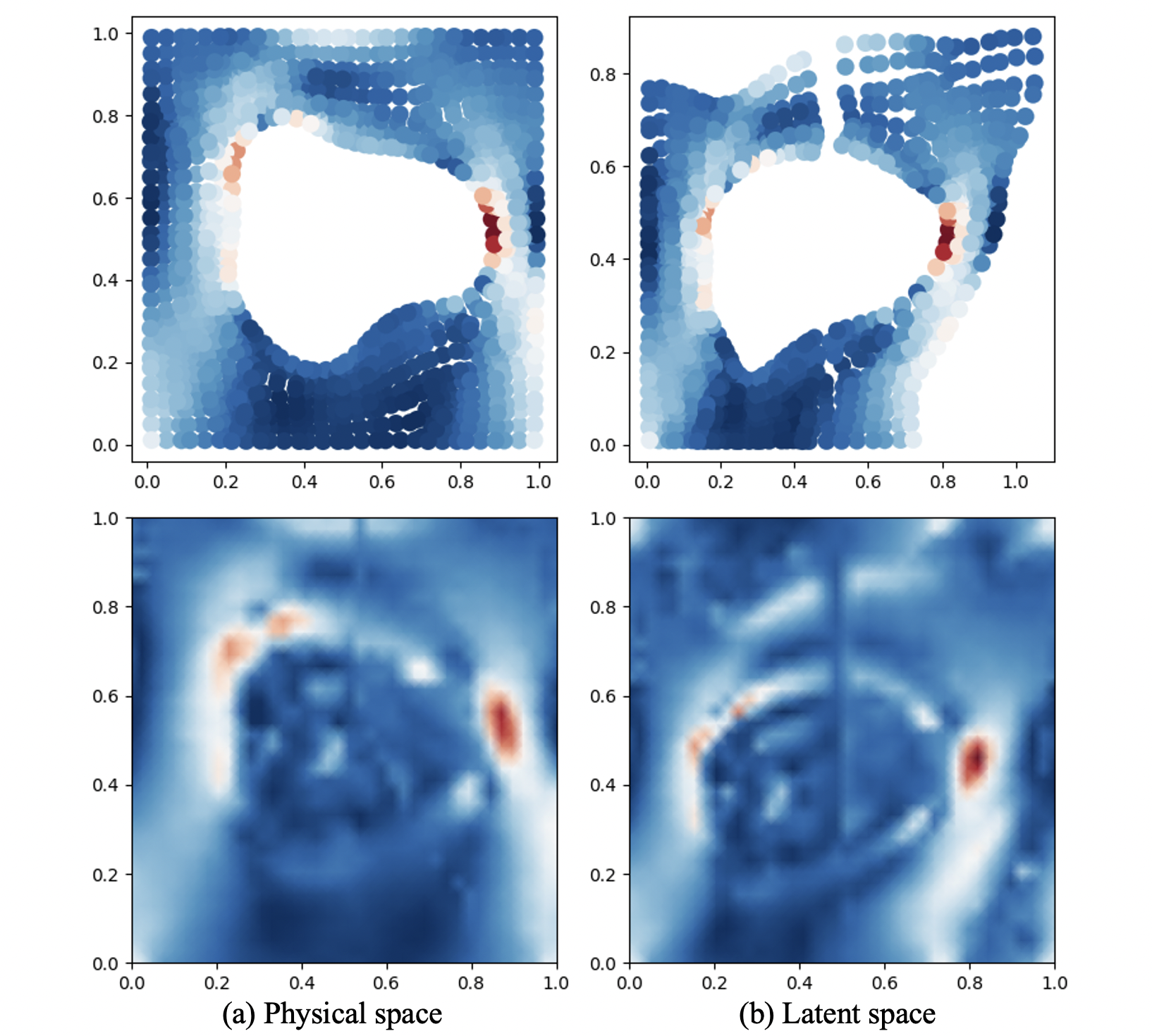}
    \caption{Visualization of $\phi_{a}^{-1}$}
    \label{fig:iphi}
    {\small The left column is the prediction of Geo-FNO on the physical space $\G_\theta(a)(x)$; the right column is the prediction of Geo-FNO on the computational space $\G_\theta(a)(\xi) = \G_\theta(a)(\phi_{a}^{-1}(x))$. The top row is the solution on the input mesh. The bottom row is the full-field solution. The latent space has a cleaner wave structure.}
\end{figure}

\begin{remark}[Visualization of $\phi_{a}^{-1}$]
Figure \ref{fig:iphi} shows the effects of the coordinate transform, where (a) is the prediction $\G_\theta(a)(x)$ on the original physical mesh and (b) is the prediction on the computational mesh (without transforming back to the physical space) $\G_\theta(a)(\phi_{a}^{-1}(x))$. 
The top row visualizes the solution on the original mesh $T^i$ and $\phi_{a}^{-1}(T^i)$, while the bottom row is the full field solution on $T^c$ directly evaluated on the Fourier coefficients.
The solution on the latent space has more "white region", which is latent encoding that does not show up in the final solution. This latent encoding is similar to the Fourier continuation, making the solution easier to be represented with the Fourier basis.
As shown in the figure, the solution function on the computational mesh has a cleaner wave structure and is more periodic compared to the physical space, allowing it to be better represented by the Fourier basis.
Interestingly, there exists a vertical discontinuity on the latent space around $x=0.5$ in the latent encoding. Since most features are horizontal, the vertical discontinuity does not affect the output. The gap can be seen as a result of encoding in the high-dimensional channel space.
\end{remark}

\subsection{Plastic problem} 
\label{ssec:solid-plastic}
We consider the plastic forging problem where a block of material $\Omega = [0,L] \times [0,H]$ is impacted by a frictionless, rigid die at time $t = 0$. The die is parameterized by an arbitrary function $S_d \in H^1([0,L];\mathbb{R})$ and traveled at a constant speed $v$. 
{The block is clamped on the bottom edge and the displacement boundary condition is imposed on the top edge.}
The governing equation is the same as the previous example but with an elasto-plastic constitutive model 
{ given by 
\begin{align*}
\bm{\sigma} &= \textbf{C} : (\bm{\epsilon}-\bm{\epsilon}_p)\\
\dot{\bm{\epsilon}}_p &= \lambda \nabla_{\bm{\sigma}} f(\bm{\sigma})\\
f(\bm{\sigma}) &= \sqrt{\frac{3}{2}}|\bm{\sigma}-\frac{1}{3}\text{tr}(\bm{\sigma})\cdot I|_F - \bm{\sigma}_Y
\end{align*}
where $\lambda$ is the plastic multiplier constrained by
$\lambda \geq 0 , f(\bm{\sigma})  \leq 0, $
and 
$\lambda \cdot f(\bm{\sigma})  = 0.$
The isotropic stiffness tensor $\mathbf{C}$ is with Young's modulus $E = 200$ GPa and Poisson's ratio 0.3.} The yield strength $\bm{\sigma}_Y$ is set to 70 MPa with the mass density $\rho^s = 7850 \text{kg}\cdot \text{m}^{-3}$.  
We generate 900 training data and 80 test data by using the commercial Finite Element software ABAQUS \cite{0b112d0e5eba4b7f9768cfe1d818872e}, using 3000 4-node quadrilateral bi-linear elements (CPS4R in ABAQUS's notation). Without lose of generality, we fix $L = 50$mm, $H = 15 $mm, and $v = 3$ ms$^{-1}$. For each sample, we generate a random function $S_d$ by randomly sampling $S_d(x_k)$ on $\{x_k = k L/7; k = 0,..,7\}$ with a uniform probability measure. The die geometry is then generated by interpolating $S_d(x_k)$ using piecewise Cubic Hermit Splines. It takes about 600 CPU seconds for each simulation. The target solution operator maps from the shape of the die to the time-dependent mesh grid and deformation. The data is given on the $101\times31$ structured mesh with 20 time steps.

\begin{figure}
    \centering
    \includegraphics[width=\textwidth]{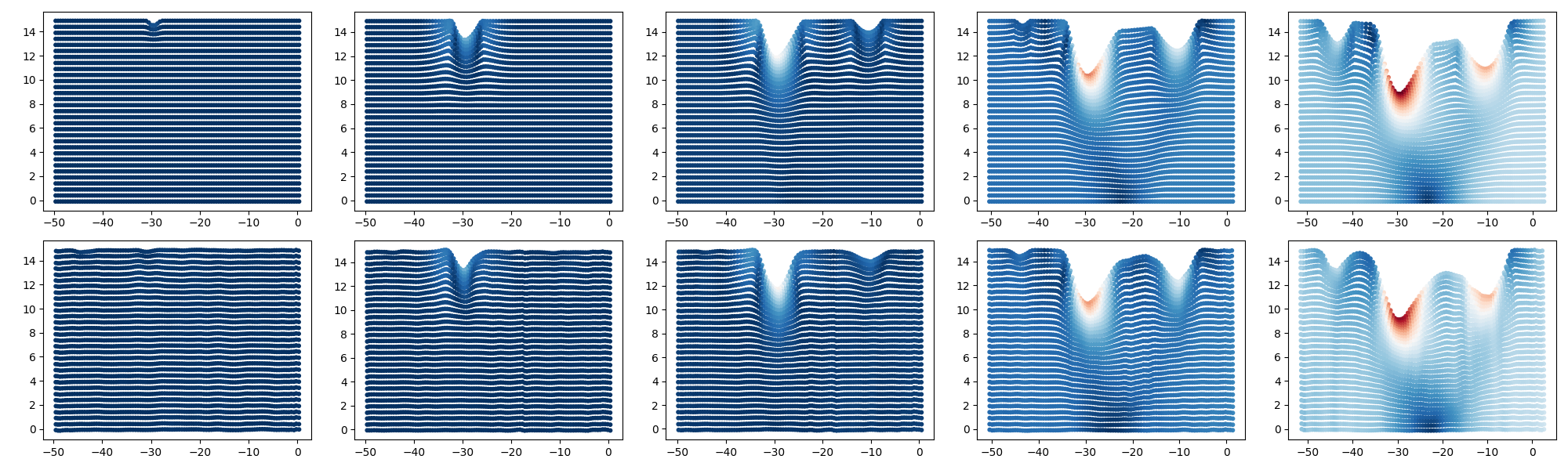}
    \caption{Geo-FNO on Plasticity}
    \label{fig:plas}
    {\small The top row is the truth and the bottom row is the prediction. The five columns represent the changes in time. The color represents the norm of displacement.}
\end{figure}

The plastic forging problem is a time-dependent problem, so we used the FNO3d model as the backbone to do convolution in both spatial dimension and temporal dimensions. 
Since the data is given on a structured mesh, the deformation \eqref{eq:structured} has an analytical form, so there is no need to learn a deformation. Geo-FNO is equivalent to directly applying the standard FNO on the structured mesh, and hence it perseveres the speed of standard FNO with an inference time of around $0.01$ second per.
In this experiment, Geo-FNO outputs both the position of the grid as well as the displacement at each grid point. As shown in \cref{fig:plas}, Geo-FNO correctly predicts the evolution of the shape and the displacement. It has a moderate error on the top of the material, which is flat in the ground truth but curved in the prediction. Overall, Geo-FNO serves as an efficient surrogate model with test error  0.0074 (See \cref{table:structured}).

\begin{table*}
\begin{center}
\begin{tabular}{c|cc|cc|cc}
\multicolumn{1}{c}{\bf Model}
&\multicolumn{2}{|c}{\bf Airfoil}
&\multicolumn{2}{|c}{\bf Pipe}
&\multicolumn{2}{|c}{\bf Plasticity}\\
& training & testing & training & testing & training & testing\\
\hline 
\hline 
Geo-FNO & 0.0134 & \textbf{0.0138} & 0.0047 & \textbf{0.0067} & 0.0071 & \textbf{0.0074} \\
FNO interpolation & $0.0287$ & $0.0421$ & $0.0083$ & $0.0151$ & $-$ & $-$  \\
UNet interpolation & $0.0039$ & $0.0519$ & $0.0109$ & $0.0182$ & $-$ & $-$ \\
\hline
\hline 
\end{tabular}\\
{\small The Plasticity requires the mesh as a target output, so interpolation methods do not apply.}
\end{center}
\caption{Benchmark on airfoils, pipe flows, and plasticity. Inputs are structured meshes.} 
\label{table:structured}
\end{table*}

\subsection{Advection equation on unit sphere}
\label{ssec:fluid-sphere}
We consider the cos bell test introduced in~\cite{rasch1994conservative}, where a cos bell
$
c(0,x) = h_c(1.0 + \cos(\pi\frac{\textrm{dist}(x, x_c)}{r}))
$
is randomly generated and centered at $x_c = (\lambda_c, \theta_c)$ with radius $r\sim\mathbb{U}[\frac{10\pi}{128},\, \frac{20\pi}{128}]$  and height $h_c\sim \mathbb{U}[0.5,\, 1.5]$ on the unit sphere. Here $\textrm{dist}$ denotes great-circle distance, $\lambda_c, \theta_c \sim \mathbb{U}[\frac{-\pi}{3}, \frac{\pi}{3}]$ denote longitude and latitude. The cos bell is advected following
$$\frac{\partial c}{\partial t} + \nabla(\bm{v} c) = 0,$$
here the advective speed $\bm{v}(\lambda, \theta) = \bigl(\cos\beta +\sin\beta \tan\theta \cos\lambda, -U\sin\beta\sin\lambda \bigr)$ with $\beta = \pi/2$ and $U = 2\pi/256$. We generate $1000$ training data and $200$ test data with a finite volume solver. 
The inputs $c(0,x)$ are given as point clouds with a size of around $8000$ on the unit sphere. The target output is the solution $c(t,x)$ at $t = 256/3$.

\begin{figure}
    \centering
    \includegraphics[width=\textwidth]{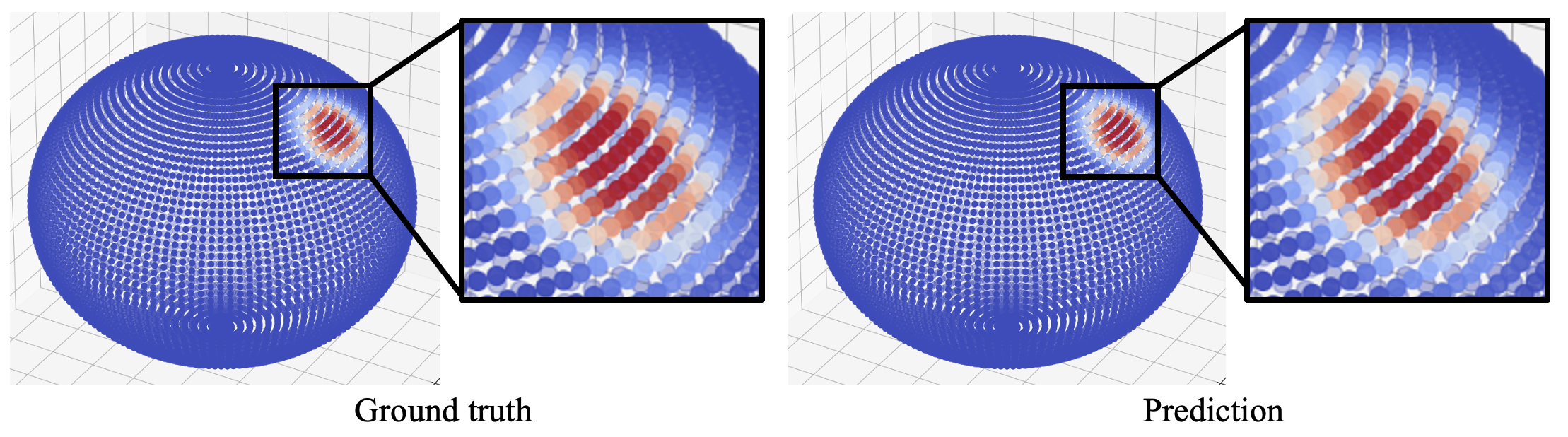}
    \caption{{Advection equation on the unit sphere. It shows that 2D Geo-FNO with domain decomposition can simulate PDE on general topologies.}}
    \label{fig:sphere}
\end{figure}

As shown in Figure \ref{fig:sphere} and table \ref{table:sphere}, Geo-FNO can work on topology differently than torus or disk. 
{Through domain decomposition, particularly by splitting the sphere into the northern and southern hemispheres, we can apply FFT2D for the spherical problem without raising dimensionality.}
It outperforms baseline models such as the original FNO and UNet models applied on the 2D spherical map projection, whose topology is not a sphere. We also compare against the FNO3D model that embeds the sphere into $R^3$. {It exhibits a remarkably high level of test error, which we attribute to the considerably larger training data and resolution demands associated with 3D learning. This underscores the significance of selecting an appropriate latent computation domain.
}

\begin{table*}
\begin{center}
\begin{tabular}{cc|rrr|cc}
\multicolumn{1}{c}{\bf Model}
&\multicolumn{1}{c}{\bf topology}
&\multicolumn{1}{c}{\bf mesh size}
&\multicolumn{1}{c}{\bf model size}
&\multicolumn{1}{c}{\bf training error }
&\multicolumn{1}{c}{\bf testing error }\\
\hline 
\hline 
FNO2D & $T^2$ & 128x64 & 2368001  &\textbf{0.0119}& {0.0381}  \\
FNO3D & $D^3$ & 64x64x64 & 35397665   &$>100\%$& $>100\%$  \\
Geo-FNO2D & $D^2 \# D^2$ & 128x64 & 4736002  &\textbf{0.0119}& \textbf{0.0332}  \\
UNet2D  & $D^2$        & 128x64 & 7752961   &0.1964 & 0.3132 \\
DeepONet  & -   & 128x64 & 2624769  &0.1113& 0.1599\\
\hline
\end{tabular}
\end{center}
\caption{{Advection equation on sphere} } 
\label{table:sphere}
\end{table*}

\subsection{Airfoil problem with Euler's equation.}
\label{ssec:fluid-airfoil}
We consider the transonic flow over an airfoil, where the governing equation is the Euler equation, as follows, 
\begin{align*}
\frac{\partial \rho^f}{\partial t} + \nabla \cdot (\rho^f \bm{v}) = 0, \quad \quad
\frac{\partial \rho^f\bm{v}}{\partial t} + \nabla \cdot (\rho^f \bm{v} \otimes \bm{v} + p \I) = 0, \quad \quad
\frac{\partial E}{\partial t} + \nabla \cdot  \Bigl( (E + p)\bm{v} \Bigr) = 0 
\end{align*}
where $\rho^f$ is the fluid density, $\bm{v}$ is the velocity vector, $p$ is the pressure, and $E$ is the total energy. The viscous effect is ignored. {The far-field boundary condition is 
$\rho_{\infty} = 1 , p_{\infty}  = 1.0 , M_{\infty} = 0.8 , AoA = 0$
where $M_{\infty}$ is the Mach number and $AoA$ is the angle of attack, and at the airfoil, no-penetration condition is imposed.} 
The shape parameterization of the airfoil follows the design element approach~\cite{farin2014curves}.
The initial NACA-0012 shape is mapped onto a `cubic' design element with $8$ control nodes, and the initial shape is morphed to a different one following the displacement field of the control nodes of the design element. The displacements of control nodes are restricted to vertical direction only with prior $d\sim \mathbb{U}[-0.05, 0.05]$.
We generate $1000$ training data and $200$ test data with a second-order implicit finite volume solver. The C-grid mesh with about ($200 \times 50$) quadrilateral elements is used, and the mesh is adapted near the airfoil but not around the shock.
It takes about $1$ CPU-hour for each simulation.
The mesh point locations and Mach number on these mesh points are used as input and output data.

\begin{figure}
    \centering
    \includegraphics[width=14cm]{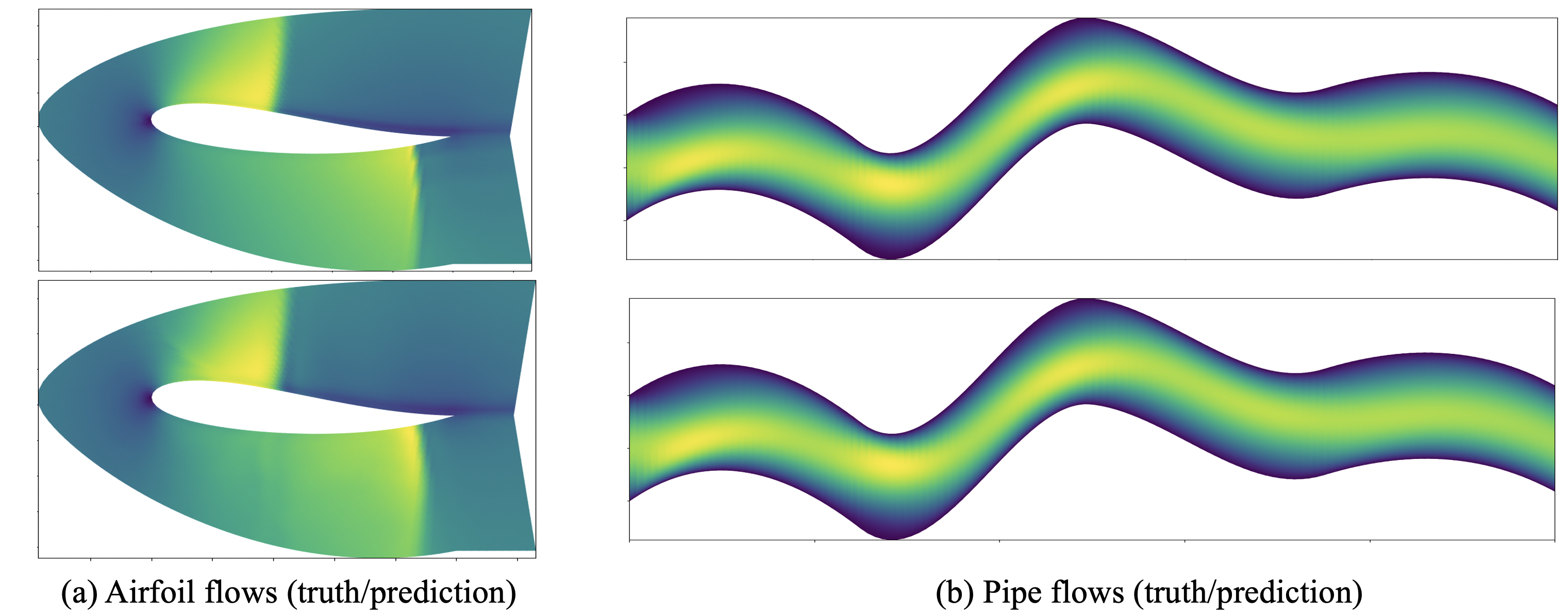}
    \caption{The airfoil flows (a) and pipe flows (b) introduced in \cref{ssec:fluid-airfoil,ssec:fluid-pipe}; The comparison is shown between the reference obtained using a traditional solver (top) and the Geo-FNO result (bottom).
    }
    \label{fig:fluid}
\end{figure}

{
Similar with the plastic problem, the data for the airfoil problem is given on a structured mesh (Omesh), the deformation \eqref{eq:structured} is prescribed, so there is no need to learn a deformation. Geo-FNO is equivalent to directly applying the standard FNO on the structured mesh.  The prediction of Geo-FNO on a sample test is presented in \cref{fig:fluid}-a, although there are two shocks induced by the airfoil, the Geo-FNO prediction matches well with the reference. We also compare 
Geo-FNO with the interpolation-based ML methods, which directly interpolates the target on a larger rectangular space. As shown in \cref{table:structured}, Geo-FNO outperforms both FNO and UNet with interpolation. 
In the following discusses, we explore discretization convergence of Geo-FNO model, throughly compare the cost-accuracy trade-off between Geo-FNO and traditional Euler solvers. And finally we demonstrate that we can conduct real-time design optimization with the trained Geo-FNO model.
}

\paragraph{Discretization Convergence}
Discretization-convergence is an important property for surrogate models \citep{kovachki2021neural}. It means the model can be applied to any discretizations and resolutions, and further, the same set of parameters can be transferred to different discretizations and resolutions. Such a design philosophy guides the research on neural operators to model the target solution operator as a mapping between function spaces, not just a specific model at one testing resolution. The graph neural operator and Fourier neural operator (with discrete Fourier transform) are both discretization-convergent. However, when implemented with the Fast Fourier transform, Fourier neural operator is restricted to the uniform grids and therefore it loses the discretization-convergent property. The proposed Geo-FNO model, on the other hand, extends FNO (with FFT) to non-uniform meshes, and retains discretization-convergence. Geo-FNO can be trained on a low-resolution dataset and evaluated at a higher resolution. On the Airfoil problem, we train Geo-FNO on a $56 \times 51$ mesh. It achieves test errors of $0.0147, 0.0329$, and $0.0428$ on $56 \times 51$, $111 \times 51$, and $221 \times 51$ meshes, respectively. Conventional deep learning methods such as U-Nets are not capable of transferring among different resolutions.

\paragraph{Cost-Accuracy Study}
Geo-FNO is used as the surrogate model to efficiently approximate expensive PDE simulations.
Such surrogate modeling is an enabling methodology for many-query computations in science and engineering, e.g., design optimization. 
In principle, the relative merits of different surrogate models can
be evaluated by understanding, for each one, the cost required to achieve a given level of accuracy.
De Hoop et. al. demonstrates in~\citep{de2022cost}, 
that FNO has a superior cost-accuracy trade-off among different neural operator approaches. 
In this section, we study the cost-accuracy trade-off in comparison with traditional numerical solvers with different resolutions. 
Specifically, we consider the Euler equation airfoil test ~\ref{ssec:fluid-airfoil}, where the reference data are generated by a $220 \times 50$ grid with $2000$ implicit backward-Euler pseudo-time-stepping iterations. 
For comparison, we use the same solver but with different spatial resolutions: $220 \times 50$, $110 \times 20$, and $44 \times 5$ grids, and different time integrators, including implicit backward-Euler and explicit Runge-Kutta-2 schemes, with different pseudo time-stepping iterations. For each setup, we estimate the error and CPU time by an average of 10 runs sampled from the data set. The cost-accuracy trade-off is depicted in Figure~\ref{fig:cost-accuracy}. Geo-FNO is at least $10^4$x faster while having the same accuracy, for the presented test. We should also mention, the speed-up of Geo-FNO is due to the avoidance of time-stepping iterations, Euler flux computation, and GPU acceleration.

\begin{figure}
    \centering
    \includegraphics[width=14cm]{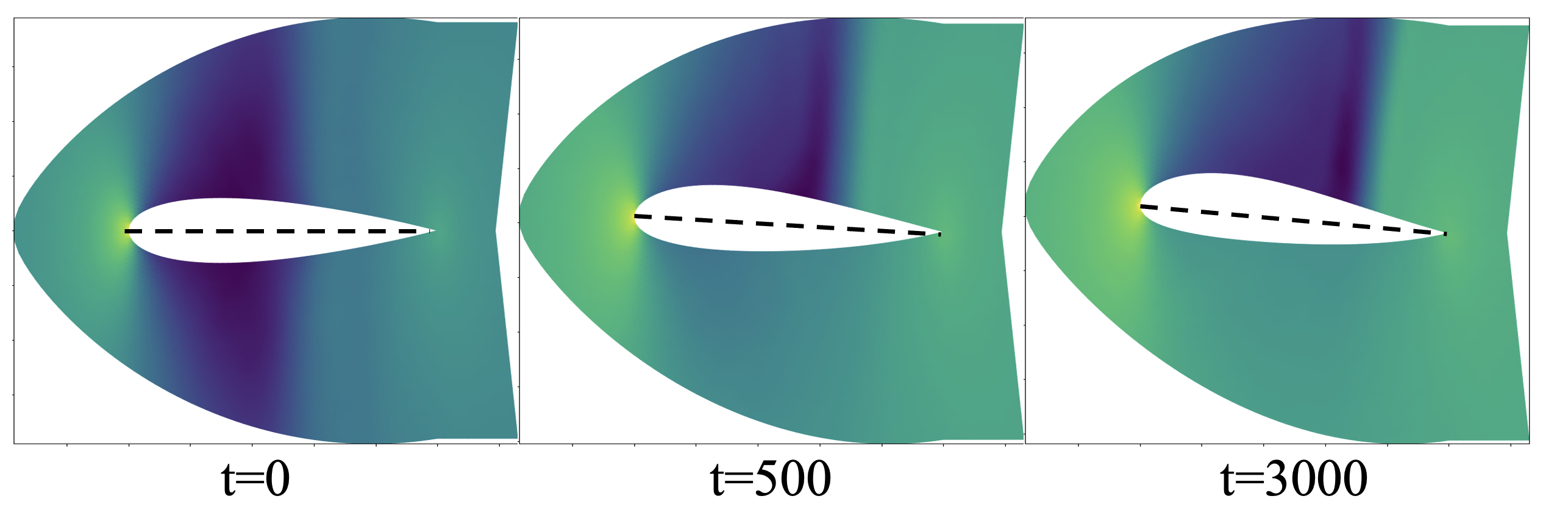}
    \caption{{The inverse design for the airfoil flow problem with end-to-end optimization. The optimal design using the simulation from the numerical solver matches the prediction from Geo-FNO.}}
    \label{fig:design}
\end{figure}

\paragraph{Inverse design} Once the Geo-FNO model is trained, it can be used to do the inverse design.  We can directly optimize the design parameters to achieve the design goal. For example, as shown in \cref{fig:design}, the shape of the airfoil is parameterized by the { vertical displacements of seven spline nodes}. We set the design goal to minimize the drag lift ratio. We first train the model mapping from the input mesh to the output pressure field, 
{then program the maps from the vertical  displacement of spline nodes to input mesh and from the output pressure field to the drag lift ratio
and finally optimize the vertical displacement of spline nodes in an end-to-end manner.} 
As shown in the figure, the resulting airfoil becomes asymmetry with larger upper camber over the optimization iteration, which increases the lift coefficient and matches the physical intuition. { Finally, we use the traditional numerical solver to verify optimized design shape. For the optimized design shape, both Geo-FNO and traditional numerical solver lead to a drag coefficient of $0.04$ and a lift coefficient of $0.29$.
}

\subsection{Pipe problem with Navier-Stokes equation}
\label{ssec:fluid-pipe}
Finally, we consider the incompressible flow in a pipe, where the governing equation is the incompressible Navier-Stokes equation, as follows, 
 \begin{align*}
\frac{\partial \bm{v}}{\partial t} + (\bm{v}  \nabla) \bm{v} =-\nabla p + \nu \nabla^2 \bm{v}, \quad \quad 
\nabla  \cdot  \bm{v} = 0
\end{align*}
where $\bm{v}$ is the velocity vector, $p$ is the pressure, and $\nu = 0.005$ is the viscosity. {The parabolic velocity profile with maximum velocity $\bm{v} = [1, 0]$ is imposed at the inlet, the free boundary condition is imposed at the outlet, and no-slip boundary condition is imposed at the pipe surface.}
The pipe is of length $10$ and width $1$, and the centerline of the pipe is parameterized by 4 piecewise cubic polynomials, which are determined by the vertical positions and slopes on $5$ spatially uniform control nodes. The vertical position at these control nodes obeys  $d\sim \mathbb{U}[-2, 2]$ and the slop at these control nodes obeys  $d\sim \mathbb{U}[-1, 1]$.
We generate $1000$ training data and $200$ test data with an implicit finite element solver with about $4000$ Taylor-Hood Q2-Q1 mixed elements~\cite{huang2020high}.
It takes about $70$ CPU seconds for each simulation.
The mesh point locations ($129\times 129$) and horizontal velocity on these mesh points are used as input and output data.

Similar with the plastic problem, the data for the pipe problem is given on a structured mesh, the deformation \cref{eq:structured} has an analytical form, so there is no need to learn a deformation. Geo-FNO is equivalent to directly applying the standard FNO on the structured mesh.
The prediction of Geo-FNO on a sample test is presented in \cref{fig:fluid}-b, the Geo-FNO is able to capture the boundary layer, and the prediction matches well with the reference. 
Geo-FNO also outperforms these interpolation-based ML ethods, which directly interpolates the target on a larger rectangular space, as shown in \cref{table:structured}. 

\section{Conclusion and future works}
\label{sec:conclusion}

In this work, we propose a geometry-aware FNO framework (Geo-FNO) that applies to arbitrary geometries and a variety of input formats.
The Geo-FNO deforms the irregular input domain into a uniform latent mesh on which the FFT can be applied. Such deformation can be fixed or learned with the FNO architecture in an end-to-end manner. The Geo-FNO combines both the computational efficiency of the FFT  and the flexibility of learned deformations. It is as fast as the original FNO but can represent the variations in solutions and domain shapes more efficiently and accurately.
In the end, we also discussed several potential extensions of Geo-FNO.

\paragraph{Physics-informed settings.}
In this work, we mainly consider learning surrogate models in the data-driven setting. However, the dataset may not always be available. In this case, we can use the physics-informed setting by optimizing the physics-informed equation loss.
Given a fixed input $a$, the output function $u = \G_\theta(a)$ can be explicitly written out using the formula \eqref{eq:G} and \eqref{eq:iFa}. The derivative of the deformed basis $\psi_a = e^{2i \pi \langle \phi_{a}^{-1}(x), k \rangle}$ can be exactly computed using chain rule with the auto-differentiation of the neural network $ \phi_{a}^{-1} $. Using the exact gradient, one can minimize the residual error $\R(\G_\theta(a))$ to find out the parameter $\G_\theta(a)$ that represents the solution function. The Geo-PINO method will be an optimization-based spectral method for general geometry.

\paragraph{General topologies.}
In this work, we mainly studied simple topologies of 2D disks or 2D disks with holes. If the problem topology is more challenging, there does not exist a diffeomorphism from the physical space to the uniform computational space. Thankfully, we can use the Fourier continuation and decomposition to convert the problem domain into simpler ones. It is known that 2D connected orientable surfaces can be classified as either a sphere or an n-genus torus. For spheres, it is natural to use the unit sphere as the computational space and the spherical harmonics as the computational basis. For n-genus torus ($n\geq 2$), usually, there do not exist useful harmonics series, but we can decompose the domain, which requires training multiple FNO models on each of the sub-domain in a coupled manner. We leave the domain decomposition as exciting future work. 

\paragraph{Theoretical guarantees.}
In the end, it will be interesting to extend the universal approximation bound of Fourier neural operator\citep{kovachki2021universal} to the solution operator of PDEs defined on general geometries. In \cite{kovachki2021universal}, the approximation is achieved by using existing pseudo-spectral solvers. Since FNO's architecture can approximate the operation in the pseudo-spectral solvers, FNO can approximate the target solution operators. For general domains, usually, there does not exist a pseudo-spectral solver. However, we can transform the problem into a computational space. By applying the universal approximation bound on the deformed equation $\R^c = 0$, as well as the approximation bound for the neural network $\phi_{a}^{-1}$, it is possible to obtain a bound for Geo-FNO. We also leave this direction as future work.

\appendix
\begin{table}[h]
\caption{table of notations}
\label{table:notations}
\begin{center}
\begin{tabular}{|l|l|}
\multicolumn{1}{c}{\bf Notation} 
&\multicolumn{1}{c}{\bf Meaning}\\
\hline
{\bf Fourier neural operator} &\\
$u \in \mathcal{U}$ & The solution function.\\
$\Gtrue$ & The target solution operator.\\
$\F, \F^{-1}$ & Fourier transformation and its inverse.\\
$R$ & The linear transformation applied on the lower Fourier modes.\\
$W$ & The linear transformation (bias term) applied on the spatial domain. \\
$k$ & Fourier modes / wave numbers.\\
\hline
{\bf Physical Domain} &\\
$a \in \mathcal{A}$  & the geometry parameters \\
$D_a$  & the physical domain. \\
$x \in D_a$  & the spatial coordinate of the physical domain. \\
$\T_a = \{x_i\}$  & the meshes of the physical domain  \\
$u, v \in \mathcal{U}(D_a)$ & functions defined on the physical domain\\
$\psi_a \in \mathcal{L}(D_a)$ & the deformed Fourier basis defined on the physical domain.\\
\hline
{\bf Computational Domain} &\\
$\phi_a$ & the coordinate transform maps from the computational domain \\& to the physical domain.\\
$D^c$  & the computational domain (unit torus). \\
$\xi \in D^c$  & the spatial coordinate of the computational domain. \\
$\T^c = \{\xi_i\}$  & the meshes of the computational domain  (a uniform mesh). \\
$u^c, v^c \in \mathcal{U}(D^c)$ & functions defined on the computational domain.\\
$\psi^c \in \mathcal{L}(D^c)$ & the standard Fourier basis defined on the computational domain.\\
\hline
\end{tabular}
\end{center}
\end{table}

\bibliography{ref}




\end{document}